\documentclass[sort,numbers]{article}
\newcommand{\cmmnt}[1]{}
\pdfoutput=1

\usepackage[final]{nips_2018}
\usepackage[toc,page]{appendix}
\usepackage[utf8]{inputenc} 
\usepackage[T1]{fontenc}    
\usepackage{hyperref}       
\usepackage{url}            
\usepackage{booktabs}       
\usepackage{amsfonts}       
\usepackage{nicefrac}       
\usepackage{microtype}      
\usepackage{times}  
\usepackage{helvet}  
\usepackage{courier}  
\usepackage{url}  
\usepackage{graphicx}  
\usepackage{booktabs}
\usepackage{subcaption}
\usepackage{xcolor}
\usepackage{color}
\usepackage{todonotes}
\usepackage{wrapfig}
\usepackage{ulem}
\usepackage{pifont}
\usepackage{enumitem}
\usepackage{amsmath}
\usepackage{comment}

\title{How2: A Large-scale Dataset for Multimodal Language Understanding}

\newcommand{\dataset}{\texttt{How2}}
\newcommand{\english}{subtitles}
\newcommand{\cmark}{\ding{51}}%
\newcommand{\xmark}{\ding{55}}%

\renewcommand{\cite}{\citep}

\author{Ramon Sanabria\\Carnegie Mellon University\\
\url{ramons@cs.cmu.edu}
\And Ozan Caglayan\\Le Mans University\And Shruti Palaskar\\Carnegie Mellon University\AND Desmond Elliott\\University of Copenhagen\And Lo\"ic Barrault\\Le Mans University\AND Lucia Specia\\University of Sheffield\And Florian Metze\\Carnegie Mellon University}

\begin{document}

\maketitle
\begin{abstract}
Human information processing is inherently multimodal, and language is best understood in a situated context. In order to achieve human-like language processing capabilities, machines should be able to jointly process multimodal data, and not just text, images, or speech in isolation. Nevertheless, there are very few multimodal datasets to support such research, resulting in a limited interaction among different research communities. In this paper, we introduce {\dataset}, a large-scale dataset of instructional videos covering a wide variety of topics across 80,000 clips (about 2,000 hours), with word-level time alignments to the ground-truth English {\english}. In addition to being multimodal, {\dataset} is multilingual: we crowdsourced Portuguese translations of the {\english}.
We present results for monomodal and multimodal baselines on several language processing tasks with interesting insights on the utility of different modalities. We hope that by making the {\dataset} dataset and baselines available we will encourage collaboration across language, speech and vision communities.
\end{abstract}
\section{Introduction}
Multimodal sensory integration is an important aspect of human concept representation, language processing and reasoning \cite{barsalou2003grounding}. From a computational perspective, major breakthroughs in natural language processing (NLP), computer vision (CV), and automatic speech recognition (ASR) have resulted in improvements in a wide range of multimodal tasks, including visual question-answering~\cite{antol2015vqa}, multimodal machine translation~\cite{specia2016shared}, visual dialogue~\cite{visdial}, and grounded ASR~\cite{palaskar_mmasr}.
Despite these advances, state-of-the-art computational models are nowhere near integrating multiple modalities as effectively as humans. This can be partially attributed to a lack of resources that are {\it pervasively} multimodal: existing datasets are typically focused on a single task, \textit{e.g.} images and text for image captioning~\cite{Chen2015}, images and text for visual-question answering \cite{antol2015vqa}, or speech and text for ASR~\cite{godfrey1992switchboard}. These datasets play a crucial role in the development of their fields, but their single-task nature limits the collective ability to develop general purpose artificial intelligence.

\begin{figure}[t]
\begin{subfigure}[c]{0.3\textwidth}
\includegraphics[width=3.5cm]{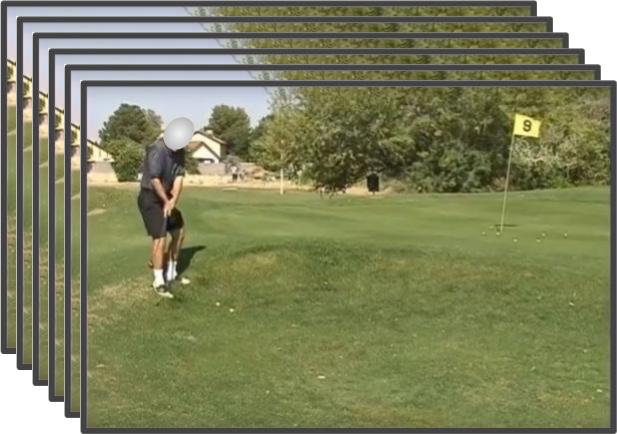}\\
\includegraphics[width=3.5cm]{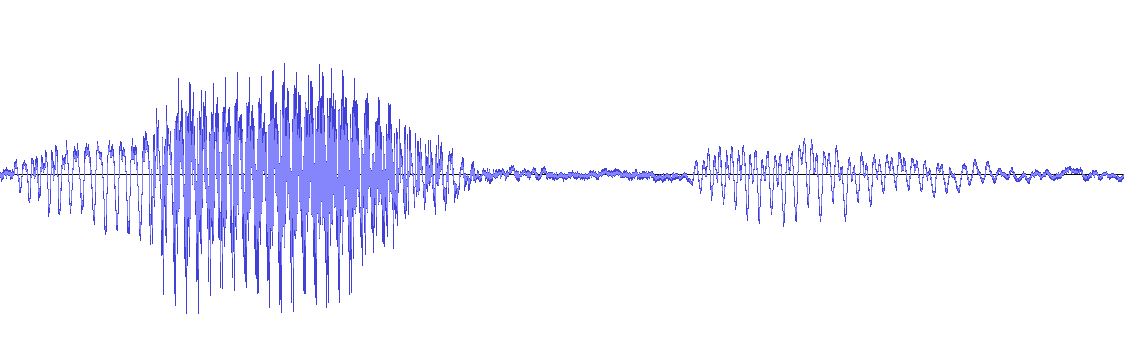}
\end{subfigure}
\begin{subfigure}[c]{0.63\textwidth}
\textbf{I'm very close to the green but I didn't get it on the green so now I'm in this grass bunker.}\\[.1cm]
\textit{Eu estou muito perto do green, mas eu n\~ao pus a bola no green, ent\~ao agora estou neste bunker de grama.}\\[.1cm]
\begin{tikzpicture}
\node [text width=25em, draw, gray, rounded corners, inner sep=1ex] {\textcolor{black}{In golf, get the body low in order to get underneath the golf ball when chipping out of thick grass from a side hill lie.} 
};
\end{tikzpicture}
\end{subfigure}
\caption{\label{fig:onecol_sample} {\dataset} contains a large variety of instructional videos with utterance-level English \english{} (in bold), aligned Portuguese translations (in italics), and video-level English summaries (in the box). Multimodality helps resolve ambiguities and improves understanding.}
\end{figure}

We introduce {\dataset}, a dataset of instructional videos paired with spoken utterances, English \english{} and their crowdsourced Portuguese translations, as well as English video summaries. The pervasive multimodality of {\dataset} makes it an ideal resource for developing new models for multimodal understanding. 
In comparison to other multimodal resources, {\dataset} is a naturally occurring dataset: neither the \english{}, nor the summaries have been crowdsourced. Furthermore, the visual content is inherently related to the spoken utterances.
Figure~\ref{fig:onecol_sample} shows an example in which the presenter is explaining how to play a golf shot. 
If one only has access to the text, 
it is unclear whether the ``{\it green}'' in the \english{} refers to the colour green (``{\it verde}'' in Portuguese), or the surface type (``{\it green}'' in Portuguese). 
The textual context alone is not enough to disambiguate the meaning of the \english{}, and at the time of writing, both Google Translate and Microsoft Translator incorrectly translate ``{\it green}'' as ``{\it verde}''. However, given additional visual context (green grass with a flag pole), or the audio context (outside with the sound of chipping a golf ball), our multimodal models can correctly interpret this utterance. See Appendix~\ref{sec:how2examples} for more examples. 

The value of additional modalities can also be demonstrated for ASR. Object and motion level visual cues can filter out systematic noise that co-occurs with activities. Scene information from an image can be used to learn a common auditory representational space for different environmental characteristics such as indoor vs. outdoor settings~\cite{miao2016open}. Entities in an image can also be used to adapt a language model towards a domain~\cite{context_mmasr}.

Together with the dataset, we also release code to perform baseline experiments on several tasks covering different subsets of {\dataset}. We find that action-level visual features improve automatic speech recognition, video summarization and speech-to-text translation. These results demonstrate the potential of the {\dataset} dataset for future multimodal research.


\section{\dataset{} Dataset}
\label{sec:dataset}
\begin{table}[t]
\centering
\caption{Statistics of {\dataset} dataset.}
\renewcommand{\arraystretch}{1.1}
\begin{tabular}{llrrrr}
\toprule
& & Videos      & Hours     & Clips/Sentences & Per Clip Statistics \\ 
\midrule
300h    & train   & 13,168  & 298.2    & 184,949 & 5.8 seconds \& 20 words \\
        & val     & 150     & 3.2      & 2,022   & 5.8 seconds \& 20 words \\
        & test    & 175     & 3.7      & 2,305   & 5.8 seconds \& 20 words \\
        & held    & 169     & 3.0      & 2,021   & 5.4 seconds \& 20 words \\
        \midrule \midrule
2000h   & train   & 73,993  & 1,766.6   & -         &                               \\
        & val     & 2,965   & 71.3      & -         &                               \\
        & test    & 2,156   & 51.7      & -         &                               \\
\bottomrule
\end{tabular}
\label{tbl:data_stats}
\end{table}

The {\dataset} dataset consists of 79,114 instructional videos (2,000 hours in total, with an average length of ~90 seconds) with English {\english}. The corpus can be (re-)created using the scripts and meta-data we made available at \url{https://github.com/srvk/how2-dataset}. The website also contains information on how to obtain pre-computed features for validation or saving computation, and how to reproduce the experimental results we present using nmtpy~\cite{nmtpy2017}. 

\paragraph{Collection} We downloaded the videos from \textit{YouTube}, along with various types of metadata, including ground-truth {\english} and summaries (referred to as ``descriptions'') in English, written by the video creators. Videos were scraped from the YouTube platform using a keyword based spider as described in~\cite{yu2014instructional}. 
In order to produce a multilingual and multimodal dataset, the English {\english} were first re-segmented into full sentences, which were then aligned to the speech at the word level. The visual features were extracted from the video \textit{clips} that correspond to these sentence-level alignments. The distribution of the duration of segments can be seen in Figure \ref{fig:histo_time}. See Appendix~\ref{sec:audio_seg} for more details on the alignment process.


To \textbf{generate translations}, we used the \textit{Figure Eight} crowdsourcing platform. After conducting a pilot experiment with a small set of languages, we chose Portuguese as target language because of the availability of workers and the quality and reliability of the annotations performed by them. 
In order to reduce the amount of time it would take to annotate the dataset, we posed translation as a post-editing task: 
in another pilot experiment, we instructed crowd workers to \textit{``choose the best translation''} from English to Portuguese among candidate translations provided by three state-of-the-art online neural machine translation systems. 
We then selected the system that was preferred most often, and had crowdworkers post-edit the candidate translations. This process is still ongoing. 


During annotation on \textit{Figure Eight}, the worker population was restricted to those living in Portugal or Brazil. Workers were paid US\$ 0.05 to watch a short video and post-edit the automatically translated Portuguese segment into correct Portuguese. Workers thus performed the annotation (and the pilots) in a multimodal setting.
To ascertain worker reliability, a content word of each 5 sentences
of the candidate translations was replaced by another random content word that was not part of the translation. If the word inserted was still present in the final translation, the annotations from that worker were discarded and the worker was banned from further contributing to the project.

At the time of writing, we had completed the collection of Portuguese translations for a \textit{300h} subset of the entire dataset from 200 workers (each was limited to 5,000 segments to post-edit but none of them reached this limit). We discarded and re-annotated 18\% of the \textit{300h}. The total cost for data collection thus far was US\$ 8,771.

In a \textbf{verification experiment}, we determined found that training an English-Portuguese neural MT system on 300h of machine generated training data degrades performance by about 1 BLEU point, when compared to a system that has been trained on the post-edited translations, when evaluated against expert-validated post-edited translations, showing that the approach is justified.

\paragraph{Topic distribution} 

We clustered the English {\english} using Latent Dirichlet Allocation (LDA) \cite{lda}. Upon analyzing the clusters with top words in each topic, inter-topic and intra-topic distances, we found that a good representation for the \textit{300h} subset consists of 22 topics. We hand-labeled these topics based on top words in each cluster, as shown in Figure~\ref{fig:topic_distrib}.

\begin{figure}
\centering
\begin{subfigure}{.5\textwidth}
  \centering
  \includegraphics[height=5cm]{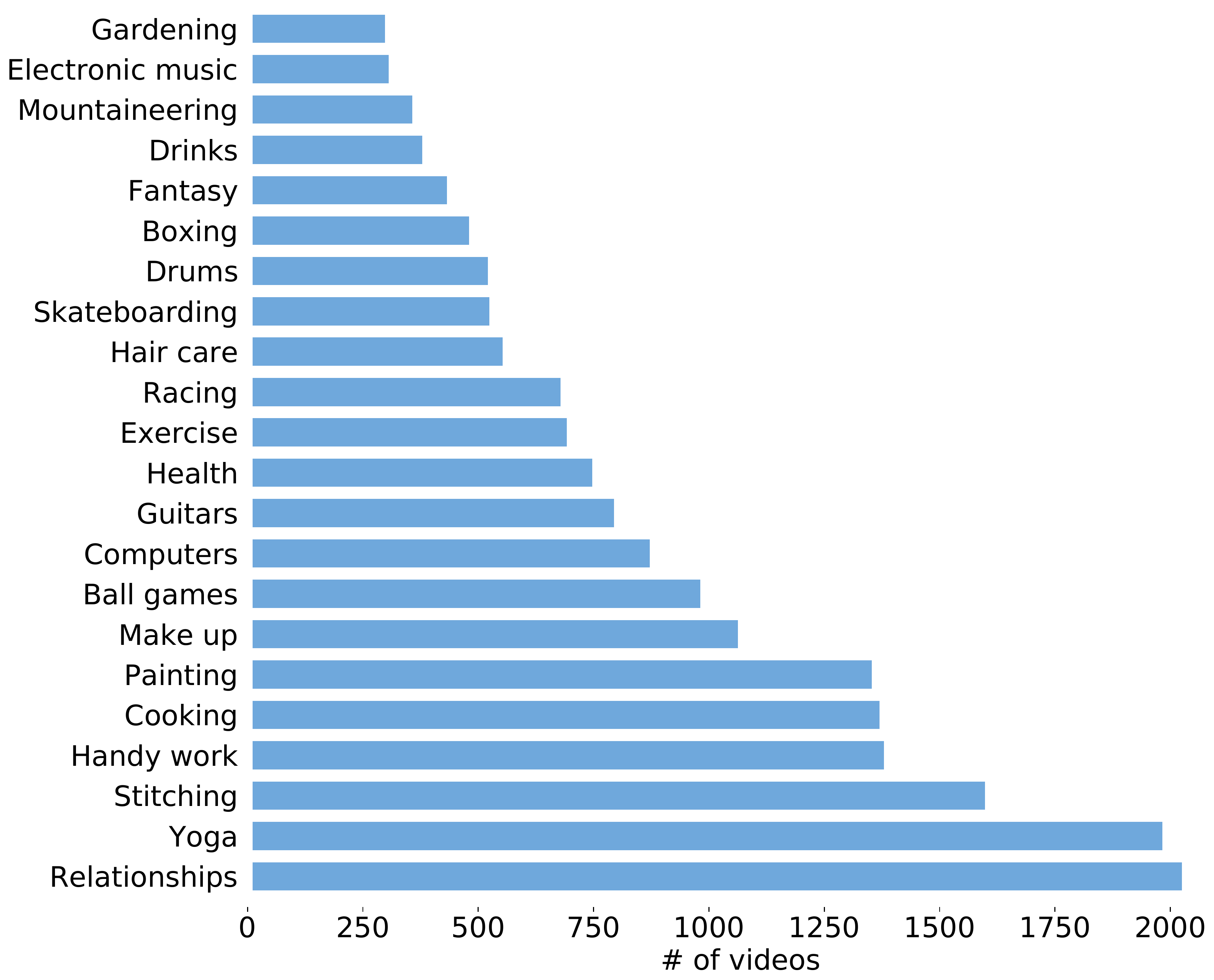}
  \caption{Topic distribution.}
  \label{fig:topic_distrib}
\end{subfigure}\hfill
\begin{subfigure}{.5\textwidth}
  \centering
  \includegraphics[height=5cm]{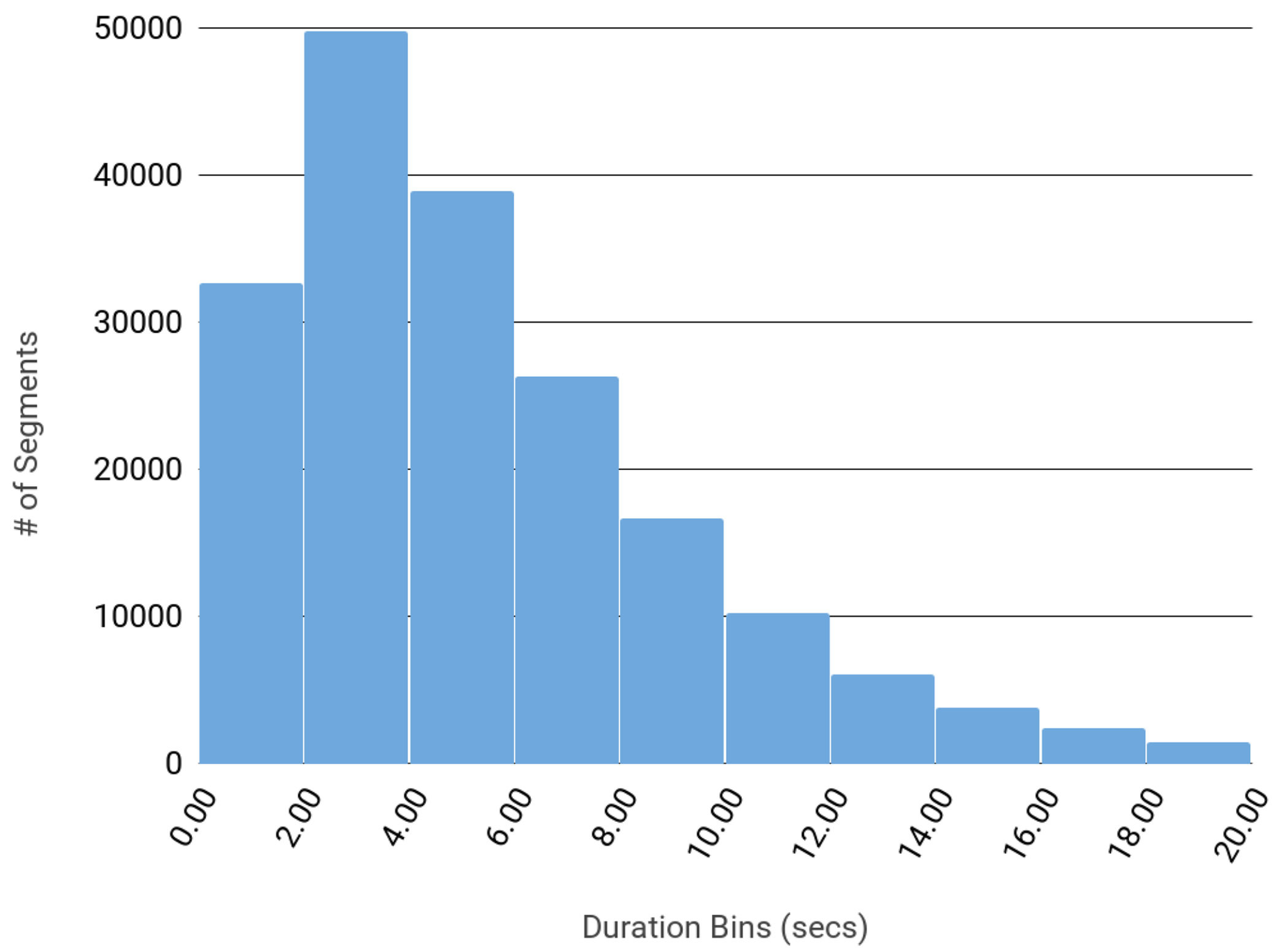}
  \caption{Segment duration.}
  \label{fig:histo_time}
\end{subfigure}
\caption{LDA topic distributions and segment duration for the \textit{300h} subset. The 2000h overall corpus exhibits very similar characteristics.}
\label{fig:how2_figures}
\end{figure}

\paragraph{Splits} Table~\ref{tbl:data_stats} presents summary statistics of the \textit{2000h} set and \textit{300h} subset: the \textit{val} and \textit{test} sets can be used for early-stopping, model selection and evaluation; the \textit{held} set is reserved for future evaluations or challenges. The total set (\textit{i.e.} 2000h) contains around 22.5M words. The tokenized training set of \textit{300h} subset contains around 3.8M (43K unique) and 3.6M (60K unique) words for English and Portuguese respectively.

\section{Experiments}
\label{sec:setup}
\begin{table}[t]
\centering
\renewcommand{\arraystretch}{1.1}
\caption{Results of the automatic speech recognition, machine translation, speech-to-text translation, and summarization experiments on \textit{test} set. The arrows indicate direction of improvement.}
\label{tbl:all_results}
\vspace{2mm}
\begin{tabular}{lcccc}
\toprule
& ASR (\% WER $\Downarrow$) & MT (BLEU $\Uparrow$) & STT (BLEU $\Uparrow$)  & SUM (ROUGE-L $\Uparrow$) \\ \midrule
Baseline                    & 19.4 & 54.4 & 36.0 & 53.9 \\
Multimodal\phantom{spc} & 18.0 & 54.4 & 37.2 & 54.9 \\
\bottomrule
\end{tabular}
\end{table}

To demonstrate and explore the potential of the \dataset{} dataset, we propose several tasks and developed systems for them using a sequence-to-sequence (S2S) approach. 
Table~\ref{tbl:all_results} summarizes the baseline results on the \textit{300h} training set for all tasks; only the  summarization task uses the entire \textit{2000h} set. More details can be found in the Appendix~\ref{sec:arch_details}.
\begin{enumerate}[leftmargin=1cm]
    \item \textbf{Automatic speech recognition.} We use an S2S model with a deep bi-directional LSTM encoder \cite{hochreiter1997long}. For multimodal ASR, we apply visual adaptive training \cite{context_mmasr,palaskar_mmasr} where we re-train an ASR model by adding a linear adaptation layer which learns a video-specific bias to additively shift the speech features. All parameters of the network all jointly trained in this re-training step. The adaptation layer increases the model size by less than 1\%.
    \item \textbf{Machine translation.} We train an S2S MT model for English$\rightarrow$Portuguese using a bidirectional GRU~\cite{cho2014gru}. For multimodal MT (MMT), we apply the same adaptive approach as we did for ASR but the inputs to be shifted are now word embeddings instead of speech features. The adaptation layer increases the model size by 8\%.
    \item \textbf{Speech-to-text translation.} We directly translate from English speech to Portuguese using the same ASR architecture but with a different target vocabulary, which is similar to previous approaches \cite{weiss2017sequence,berard2018end}. For multimodal STT, we apply the same adaptive scheme as in ASR.
    \item \textbf{Summarization.} The baseline is again the same S2S MT model. For multimodal summarization, we follow the hierarchical attention approach \cite{libovicky2017attention,libovicky2018multimodal} to combine textual and visual modalities by using a sequence of action-level features instead of an average-pooled one as in the other experiments. This latter increases the model size by 14\%.
\end{enumerate}

\section{Related work}
\label{sec:related}
\begin{table}[pt]
\centering
\caption{Comparison with previous datasets: (IC) and (VD) stand for image and video captioning. Language names are encoded in ISO-639-1.}
\label{tbl:datacomp}
\vspace{2mm}
\renewcommand{\arraystretch}{1.2}
\resizebox{.99\textwidth}{!}{
\begin{tabular}{@{}lllccl@{}}
\toprule
Task \phantom{spacing} & Dataset & Languages        & Audio  & Visual & Size                     \\ \midrule
IC & Flickr8K \cite{Hodosh2013}      & EN, TR \cite{unal2016tasviret}, ZH \cite{li2016adding} & \xmark & \cmark\,(I) & 8K  \\
IC & Flickr30K \cite{flickr30k}      & 150K EN, DE \cite{Elliott2016}  & \xmark & \cmark\,(I) & 30K    \\
IC & MSCOCO \cite{Chen2015}          & 414K EN                           & \xmark & \cmark\,(I) & 82K \\
IC &                            & 820K JA \cite{yoshikawa2017stair} & \xmark & \cmark\,(I) & 164K \\ \midrule
MMT & Multi30K \cite{Elliott2016}             &     EN, DE, FR \cite{elliott2017shared}, CZ \cite{barrault2018shared}                          & \xmark & \cmark\,(I) & 30K  \\ \midrule
VD & MSVD \cite{ChenDollan2011}              & 122K total in many              & \cmark & \cmark\,(V) & \cmmnt{2K clips} 5.3 hours \\
VD & LSMDC \cite{lsmdc}                      & EN                              & \cmark & \cmark\,(V) & \cmmnt{128K clips} 150 hours \\
VD & MSRVTT \cite{Xuetal:2016_MSR-VTT}       & EN                              & \cmark & \cmark\,(V) &  \cmmnt{1200K clips} 41 hours \\ \midrule
AV-ASR & Grid \cite{grid}                        & EN                             & \cmark & \cmark\,(V) & 50 hours \\
AV-ASR & ViaVoice \cite{neti2000audio}           & EN                             & \cmark & \cmark\,(V) & 34.9 hours \\
AV-ASR & LRW \cite{Chung16}                      & EN                             & \cmark & \cmark\,(V) &   800 hours  \\ \midrule
STT &  Fisher \cite{fishercallhome}              & EN, ES                           & \cmark & \xmark &   150 hours\\
STT &  Audiobooks \cite{KOCABIYIKOGLU18.621}          & EN, FR                           & \cmark & \xmark &   236 hours\\ \midrule
SUM & CNN/DMC \cite{cnndm,cnndm_nallapathi}    & EN                             & \xmark & \xmark &  286,817 pairs \\ 
SUM & DUC \cite{over2007duc}                      & EN                            & \xmark & \xmark & 500 pairs \\
SUM & \cite{li2017multi}                     & EN CZ                           & \cmark  & \cmark (V) & 492,402 pairs \\
\bottomrule
\end{tabular}}
\label{tab:datacomp}
\end{table}

Lying at the intersection of NLP and CV~\cite{Bernardi2016}, image captioning (IC) is the multimodal task with the largest number of datasets  available. The most widely used datasets in this field are the ones with human crowdsourced descriptions, such as Flickr8K~\cite{Hodosh2013}, Flickr30K~\cite{Young2014}, MSCOCO~\cite{Chen2015} and their extensions to other languages. A closely related task to IC is multimodal machine translation. So far, MMT has been addressed using captioning datasets extended with translations in different languages such as IAPR-TC12~\cite{Grubinger2006} and Multi30K which is an extension of Flickr30K into German \cite{Elliott2016}, French \cite{elliott2017shared}, and Czech \cite{barrault2018shared}. One major pitfall of these datasets is that they lack syntactic and semantic diversity. 

A similar task to IC is that of automatically describing videos (VD). The most popular datasets for VD are MSR-VTT~\cite{Xuetal:2016_MSR-VTT}, LSMDC~\cite{lsmdc}, and Microsoft Research Video Description (MSVD) corpus~\cite{ChenDollan2011} which is the only multilingual resource of this type providing 122K crowdsourced descriptions. However, two-thirds of the descriptions are in English and the ones in other languages are not parallel. {\dataset} offers a larger amount of data, all of which is in two languages.

Lipreading can be seen as a form of multimodal ASR, albeit not fusing information at the semantic level. Popular and large-scale datasets include
Grid~\cite{grid} and 
Lip Reading in The Wild~\cite{Chung16}.
{\dataset} is the first dataset that allows to perform multimodal ASR, using images as acoustic and linguistic context to improve accuracy. {\dataset} is also a valuable resource for speech-to-text translation, which is otherwise often performed using Fisher-Callhome \cite{fishercallhome} and Audiobooks \cite{KOCABIYIKOGLU18.621}. {\dataset} is the only corpus for multimodal STT currently available.


Multimodal neural abstractive summarization is an emerging field for which there are no well established benchmarking datasets yet. 
\citet{li2017multi} collected a multimodal corpus of news articles containing 500 videos of English news articles paired with human annotated summaries. \citet{uzzaman2011multimodal} 
collected a corpus of images, structured text and simplified compressed text for summarization of complex sentences.  More traditional text-based summarization is commonly based on CNN/Daily Mail \cite{cnndm,cnndm_nallapathi}, Gigaword \cite{gigaword_napoles2012annotated} and the Document Understanding Conference challenge data \cite{over2007duc}. 
An older version and non-released version of \dataset{} was used for experiments on learning action examples in videos in~\cite{yu2014instructional}.

\section{Conclusions}
We have introduced {\dataset}, a multimodal collection of instructional videos with English subtitles and crowdsourced Portuguese translations. We have also presented sequence-to-sequence baselines for machine translation, automatic speech recognition, spoken language translation, and multimodal summarization. By making available data and code for several multimodal natural language tasks, we hope to stimulate more research on these and similar challenges to obtain a deeper understanding of multimodality in language processing.


\section*{Acknowledgements}

This work was mostly conducted at the 2018 Frederick Jelinek Memorial Summer Workshop on Speech and Language Technologies,\footnote{\url{https://www.clsp.jhu.edu/workshops/18-workshop/}} hosted and sponsored by Johns Hopkins University. 
Lucia Specia received funding from the MultiMT (H2020 ERC Starting Grant No. 678017) and MMVC (Newton Fund Institutional Links Grant, ID 352343575) projects.
Loïc Barrault and Ozan Caglayan received funding from the CHISTERA M2CR (No. ANR-15-CHR2-0006-01).

\bibliographystyle{unsrtnat}
\bibliography{sanabria.bib}

\begin{thebibliography}{55}
\providecommand{\natexlab}[1]{#1}
\providecommand{\url}[1]{\texttt{#1}}
\expandafter\ifx\csname urlstyle\endcsname\relax
  \providecommand{\doi}[1]{doi: #1}\else
  \providecommand{\doi}{doi: \begingroup \urlstyle{rm}\Url}\fi

\bibitem[Barsalou et~al.(2003)Barsalou, Simmons, Barbey, and
  Wilson]{barsalou2003grounding}
Lawrence~W Barsalou, W~Kyle Simmons, Aron~K Barbey, and Christine~D Wilson.
\newblock Grounding conceptual knowledge in modality-specific systems.
\newblock \emph{Trends in Cognitive Sciences}, 2003.

\bibitem[Antol et~al.(2015)Antol, Agrawal, Lu, Mitchell, Batra,
  Lawrence~Zitnick, and Parikh]{antol2015vqa}
Stanislaw Antol, Aishwarya Agrawal, Jiasen Lu, Margaret Mitchell, Dhruv Batra,
  C~Lawrence~Zitnick, and Devi Parikh.
\newblock {VQA}: {V}isual {Q}uestion {A}nswering.
\newblock In \emph{Proceedings of the International Conference on Computer
  Vision (ICCV)}. IEEE, 2015.

\bibitem[Specia et~al.(2016)Specia, Frank, Sima'an, and
  Elliott]{specia2016shared}
Lucia Specia, Stella Frank, Khalil Sima'an, and Desmond Elliott.
\newblock A shared task on multimodal machine translation and crosslingual
  image description ({WMT}).
\newblock In \emph{Proceedings of the First Conference on Machine Translation
  (WMT)}. ACL, 2016.

\bibitem[Das et~al.(2017)Das, Kottur, Gupta, Singh, Yadav, Moura, Parikh, and
  Batra]{visdial}
Abhishek Das, Satwik Kottur, Khushi Gupta, Avi Singh, Deshraj Yadav,
  Jos\'e~M.F. Moura, Devi Parikh, and Dhruv Batra.
\newblock {V}isual {D}ialog.
\newblock In \emph{Proceedings of the Conference on Computer Vision and Pattern
  Recognition (CVPR)}. IEEE, 2017.

\bibitem[Palaskar et~al.(2018)Palaskar, Sanabria, and Metze]{palaskar_mmasr}
Shruti Palaskar, Ramon Sanabria, and Florian Metze.
\newblock End-to-end multimodal speech recognition.
\newblock In \emph{Proceedings of the International Conference on Acoustics,
  Speech and Signal Processing (ICASSP)}. IEEE, 2018.

\bibitem[Chen et~al.(2015)Chen, Fang, Lin, Vedantam, Gupta, Doll{\'{a}}r, and
  Zitnick]{Chen2015}
Xinlei Chen, Hao Fang, Tsung{-}Yi Lin, Ramakrishna Vedantam, Saurabh Gupta,
  Piotr Doll{\'{a}}r, and C.~Lawrence Zitnick.
\newblock Microsoft {COCO} captions: Data collection and evaluation server.
\newblock \emph{Computing Research Repository (CoRR)}, 2015.

\bibitem[Godfrey et~al.(1992)Godfrey, Holliman, and
  McDaniel]{godfrey1992switchboard}
John~J Godfrey, Edward~C Holliman, and Jane McDaniel.
\newblock Switchboard: Telephone speech corpus for research and development.
\newblock In \emph{Proceedings of the International Conference on Acoustics,
  Speech and Signal Processing (ICASSP)}. IEEE, 1992.

\bibitem[Miao and Metze(2016)]{miao2016open}
Yajie Miao and Florian Metze.
\newblock Open-domain audio-visual speech recognition: {A} deep learning
  approach.
\newblock In \emph{Proceedings of Interspeech}. ISCA, 2016.

\bibitem[Gupta et~al.(2017)Gupta, Miao, Neves, and Metze]{context_mmasr}
Abhinav Gupta, Yajie Miao, Leonardo Neves, and Florian Metze.
\newblock Visual features for context-aware speech recognition.
\newblock In \emph{Proceedings of the International Conference on Acoustics,
  Speech and Signal Processing (ICASSP)}, 2017.

\bibitem[Caglayan et~al.(2017)Caglayan, Garc\'{i}a-Mart\'{i}nez, Bardet,
  Aransa, Bougares, and Barrault]{nmtpy2017}
Ozan Caglayan, Mercedes Garc\'{i}a-Mart\'{i}nez, Adrien Bardet, Walid Aransa,
  Fethi Bougares, and Lo\"{i}c Barrault.
\newblock {NMTPY}: A flexible toolkit for advanced neural machine translation
  systems.
\newblock \emph{The Prague Bulletin of Mathematical Linguistics}, 2017.

\bibitem[Yu et~al.(2014)Yu, Jiang, and Hauptmann]{yu2014instructional}
Shoou-I Yu, Lu~Jiang, and Alexander Hauptmann.
\newblock Instructional videos for unsupervised harvesting and learning of
  action examples.
\newblock In \emph{Proceedings of the International Multimedia Conference
  (ACMM)}. ACM, 2014.

\bibitem[Blei et~al.(2003)Blei, Ng, and Jordan]{lda}
David~M Blei, Andrew~Y Ng, and Michael~I Jordan.
\newblock Latent dirichlet allocation.
\newblock \emph{Journal of Machine Learning Research (JMLR)}, 2003.

\bibitem[Hochreiter and Schmidhuber()]{hochreiter1997long}
Sepp Hochreiter and J{\"u}rgen Schmidhuber.
\newblock Long short-term memory.
\newblock \emph{Neural computation}.

\bibitem[Cho et~al.(2014)Cho, van Merrienboer, Gulcehre, Bahdanau, Bougares,
  Schwenk, and Bengio]{cho2014gru}
Kyunghyun Cho, Bart van Merrienboer, Caglar Gulcehre, Dzmitry Bahdanau, Fethi
  Bougares, Holger Schwenk, and Yoshua Bengio.
\newblock Learning phrase representations using rnn encoder--decoder for
  statistical machine translation.
\newblock In \emph{Proceedings of the Conference on Empirical Methods in
  Natural Language Processing (EMNLP)}. ACL, 2014.

\bibitem[Weiss et~al.(2017)Weiss, Chorowski, Jaitly, Wu, and
  Chen]{weiss2017sequence}
Ron~J Weiss, Jan Chorowski, Navdeep Jaitly, Yonghui Wu, and Zhifeng Chen.
\newblock Sequence-to-sequence models can directly translate foreign speech.
\newblock In \emph{Proceedings of Interspeech}. ISCA, 2017.

\bibitem[B{\'e}rard et~al.(2018)B{\'e}rard, Besacier, Kocabiyikoglu, and
  Pietquin]{berard2018end}
Alexandre B{\'e}rard, Laurent Besacier, Ali~Can Kocabiyikoglu, and Olivier
  Pietquin.
\newblock End-to-end automatic speech translation of audiobooks.
\newblock In \emph{Proceedings of the International Conference on Acoustics,
  Speech and Signal Processing (ICASSP)}. IEEE, 2018.

\bibitem[Libovick\'{y} and Helcl(2017)]{libovicky2017attention}
Jind\v{r}ich Libovick\'{y} and Jind\v{r}ich Helcl.
\newblock Attention strategies for multi-source sequence-to-sequence learning.
\newblock In \emph{Proceedings Annual Meeting of the Association for
  Computational Linguistics (ACL)}. ACL, 2017.

\bibitem[Libovick\'{y} et~al.(2018)Libovick\'{y}, Palaskar, Gella, and
  Metze]{libovicky2018multimodal}
Jind\v{r}ich Libovick\'{y}, Shruti Palaskar, Spandana Gella, and Florian Metze.
\newblock Multimodal abstractive summarization of open-domain videos.
\newblock In \emph{Proceedings of the Workshop on Visually Grounded Interaction
  and Language (ViGIL)}. NIPS, 2018.

\bibitem[Hodosh et~al.(2013)Hodosh, Young, and Hockenmaier]{Hodosh2013}
Micah Hodosh, Peter Young, and Julia Hockenmaier.
\newblock {Framing Image Description as a Ranking Task: Data, Models and
  Evaluation Metrics}.
\newblock \emph{Journal of Artificial Intelligence Research (JAIR)}, 2013.

\bibitem[Unal et~al.(2016)Unal, Citamak, Yagcioglu, Erdem, Erdem, Cinbis, and
  Cakici]{unal2016tasviret}
Mesut~Erhan Unal, Begum Citamak, Semih Yagcioglu, Aykut Erdem, Erkut Erdem,
  Nazli~Ikizler Cinbis, and Ruket Cakici.
\newblock Tasviret: G\"or\"unt\"ulerden otomatik t\"urk\c{c}e a\c{c}{\i}klama
  olu\c{s}turma \.{I}\c{c}in bir denekta\c{c}{\i} veri k\"umesi ({TasvirEt}: A
  benchmark dataset for automatic {Turkish} description generation from
  images).
\newblock In \emph{Proceesdings of the Sinyal \.{I}\c{s}leme ve
  \.{I}leti\c{s}im Uygulamalar{\i} Kurultay{\i} (SIU 2016)}. IEEE, 2016.

\bibitem[Li et~al.(2016)Li, Lan, Dong, and Liu]{li2016adding}
Xirong Li, Weiyu Lan, Jianfeng Dong, and Hailong Liu.
\newblock Adding {Chinese} captions to images.
\newblock In \emph{Proceedings of the International Conference on Multimedia
  Retrieval (ICMR)}. ACM, 2016.

\bibitem[Plummer et~al.(2017)Plummer, Wang, Cervantes, Caicedo, Hockenmaier,
  and Lazebnik]{flickr30k}
Bryan~A. Plummer, Liwei Wang, Chris~M. Cervantes, Juan~C. Caicedo, Julia
  Hockenmaier, and Svetlana Lazebnik.
\newblock Flickr30k entities: Collecting region-to-phrase correspondences for
  richer image-to-sentence models.
\newblock \emph{International Journal of Computer Vision}, 2017.

\bibitem[Elliott et~al.(2016)Elliott, Frank, Sima'an, and Specia]{Elliott2016}
Desmond Elliott, Stella Frank, Khalil Sima'an, and Lucia Specia.
\newblock Multi30k: Multilingual english-german image descriptions.
\newblock In \emph{Proceedings of the Workshop on Vision and Language}. ACL,
  2016.

\bibitem[Yoshikawa et~al.(2017)Yoshikawa, Shigeto, and
  Takeuchi]{yoshikawa2017stair}
Yuya Yoshikawa, Yutaro Shigeto, and Akikazu Takeuchi.
\newblock {STAIR} captions: Constructing a large-scale {J}apanese image caption
  dataset.
\newblock In \emph{Proceedings of the Annual Meeting of the Association for
  Computational Linguistics (ACL)}. ACL, 2017.

\bibitem[Elliott et~al.(2018)Elliott, Frank, Barrault, Bougares, and
  Specia]{elliott2017shared}
Desmond Elliott, Stella Frank, Lo\"{i}c Barrault, Fethi Bougares, and Lucia
  Specia.
\newblock Findings of the second shared task on multimodal machine translation
  and multilingual image description.
\newblock In \emph{Proceedings of the Second Conference on Machine Translation
  (WMT)}. ACL, 2018.

\bibitem[Barrault et~al.(2018)Barrault, Bougares, Specia, Lala, Elliott, and
  Frank]{barrault2018shared}
Lo\"{i}c Barrault, Fethi Bougares, Lucia Specia, Chiraag Lala, Desmond Elliott,
  and Stella. Frank.
\newblock Findings of the shared task on multimodal machine translation
  {(WMT)}.
\newblock In \emph{Proceedings of Conference on Machine Translation (WMT)}.
  ACL, 2018.

\bibitem[Chen and Dolan(2011)]{ChenDollan2011}
David~L. Chen and William~B. Dolan.
\newblock Building a persistent workforce on {Mechanical Turk} for multilingual
  data collection.
\newblock In \emph{Proceedings of The 3rd Human Computation Workshop (HCOMP)}.
  AAAI, 2011.

\bibitem[Rohrbach et~al.(2017)Rohrbach, Torabi, Rohrbach, Tandon, Pal,
  Larochelle, Courville, and Schiele]{lsmdc}
Anna Rohrbach, Atousa Torabi, Marcus Rohrbach, Niket Tandon, Chris Pal, Hugo
  Larochelle, Aaron Courville, and Bernt Schiele.
\newblock Movie description.
\newblock \emph{International Journal of Computer Vision}, 2017.

\bibitem[Xu et~al.(2016)Xu, Mei, Yao, and Rui]{Xuetal:2016_MSR-VTT}
Jun Xu, Tao Mei, Ting Yao, and Yong Rui.
\newblock {MSR-VTT:} a large video description dataset for bridging video and
  language.
\newblock In \emph{Proceedings of the International Conference on Computer
  Vision and Pattern Recognition (CVPR)}. IEEE, 2016.

\bibitem[Cooke et~al.(2006)Cooke, Barker, Cunningham, and Shao]{grid}
Martin Cooke, Jon Barker, Stuart Cunningham, and Xu~Shao.
\newblock An audio-visual corpus for speech perception and automatic speech
  recognition.
\newblock \emph{The Journal of the Acoustical Society of America}, 2006.

\bibitem[Neti et~al.(2000)Neti, Potamianos, Luettin, Matthews, Glotin, Vergyri,
  Sison, and Mashari]{neti2000audio}
Chalapathy Neti, Gerasimos Potamianos, Juergen Luettin, Iain Matthews, Herve
  Glotin, Dimitra Vergyri, June Sison, and Azad Mashari.
\newblock Audio visual speech recognition.
\newblock Technical report, IDIAP, 2000.

\bibitem[Chung and Zisserman(2016)]{Chung16}
Joon~Son Chung and Andrew Zisserman.
\newblock Lip reading in the wild.
\newblock In \emph{Proceedings of the Asian Conference on Computer Vision
  (ACCV)}. Springer, 2016.

\bibitem[Post et~al.(2013)Post, Kumar, Lopez, Karakos, Callison-Burch, and
  Khudanpur]{fishercallhome}
Matt Post, Gaurav Kumar, Adam Lopez, Damianos Karakos, Chris Callison-Burch,
  and Sanjeev Khudanpur.
\newblock Improved {Speech-to-Text} translation with the {F}isher and
  {C}allhome {Spanish-English} speech translation corpus.
\newblock In \emph{Proceedings International Workshop on Spoken Language
  Translation (IWSLT)}. ACL, 2013.

\bibitem[Kocabiyikoglu et~al.(2018)Kocabiyikoglu, Besacier, and
  Kraif]{KOCABIYIKOGLU18.621}
Ali~Can Kocabiyikoglu, Laurent Besacier, and Olivier Kraif.
\newblock {Augmenting Librispeech with French Translations: {A} Multimodal
  Corpus for Direct Speech Translation Evaluation}.
\newblock In \emph{Proceedings of the International Conference on Language
  Resources and Evaluation (LREC)}. ELRA, 2018.

\bibitem[Hermann et~al.(2015)Hermann, Kocisky, Grefenstette, Espeholt, Kay,
  Suleyman, and Blunsom]{cnndm}
Karl~Moritz Hermann, Tomas Kocisky, Edward Grefenstette, Lasse Espeholt, Will
  Kay, Mustafa Suleyman, and Phil Blunsom.
\newblock Teaching machines to read and comprehend.
\newblock In \emph{Proceedings of the International Conference on Neural
  Information Processing Systems (NIPS)}. NIPS, 2015.

\bibitem[Nallapati et~al.(2016)Nallapati, Zhou, dos Santos, Çaglar G~ulçehre,
  and Xiang]{cnndm_nallapathi}
Ramesh Nallapati, Bowen Zhou, Cicero dos Santos, Çaglar G~ulçehre, and Bing
  Xiang.
\newblock Abstractive text summarization using sequence-to-sequence rnns and
  beyond.
\newblock In \emph{Proceedings of the Conference on Computational Natural
  Language Learning (CoNLL)}. ACL, 2016.

\bibitem[Over et~al.(2007)Over, Dang, and Harman]{over2007duc}
Paul Over, Hoa Dang, and Donna Harman.
\newblock Duc in context.
\newblock \emph{Information Processing \& Management}, 2007.

\bibitem[Li et~al.(2017)Li, Zhu, Ma, Zhang, and Zong]{li2017multi}
Haoran Li, Junnan Zhu, Cong Ma, Jiajun Zhang, and Chengqing Zong.
\newblock Multi-modal summarization for asynchronous collection of text, image,
  audio and video.
\newblock In \emph{Proceedings of the Conference on Empirical Methods in
  Natural Language Processing (EMNLP)}. ACL, 2017.

\bibitem[Bernardi et~al.(2016)Bernardi, Cakici, Elliott, Erdem, Erdem,
  Ikizler-Cinbis, Keller, Muscat, and Plank]{Bernardi2016}
Raffaella Bernardi, Ruken Cakici, Desmond Elliott, Aykut Erdem, Erkut Erdem,
  Nazli Ikizler-Cinbis, Frank Keller, Adrian Muscat, and Barbara Plank.
\newblock Automatic description generation from images: A survey of models,
  datasets, and evaluation measures.
\newblock \emph{Journal of Artificial Intelligence Research (JAIR)}, 2016.

\bibitem[Young et~al.(2014)Young, Lai, Hodosh, and Hockenmaier]{Young2014}
Peter Young, Alice Lai, Micha Hodosh, and Julia Hockenmaier.
\newblock From image descriptions to visual denotations: New similarity metrics
  for semantic inference over event descriptions.
\newblock \emph{Transactions of the Association for Computational Linguistics
  (TACL)}, 2014.

\bibitem[Grubinger et~al.(2006)Grubinger, Clough, Muller, and
  Desealers]{Grubinger2006}
Michael Grubinger, Paul~D. Clough, Henning Muller, and Thomas Desealers.
\newblock The {IAPR TC-12} benchmark: A new evaluation resource for visual
  information systems.
\newblock In \emph{Proceedings of International Conference on Language
  Resources and Evaluation (LREC)}. ELRA, 2006.

\bibitem[UzZaman et~al.(2011)UzZaman, Bigham, and Allen]{uzzaman2011multimodal}
Naushad UzZaman, Jeffrey~P Bigham, and James~F Allen.
\newblock Multimodal summarization of complex sentences.
\newblock In \emph{Proceedings International Conference on Intelligent User
  Interfaces (IUI)}. ACM, 2011.

\bibitem[Napoles et~al.(2012)Napoles, Gormley, and
  Van~Durme]{gigaword_napoles2012annotated}
Courtney Napoles, Matthew Gormley, and Benjamin Van~Durme.
\newblock Annotated gigaword.
\newblock In \emph{Proceedings of the Joint Workshop on Automatic Knowledge
  Base Construction and Web-scale Knowledge Extraction}. ACL, 2012.

\bibitem[Povey et~al.(2011)Povey, Ghoshal, Boulianne, Burget, Glembek, Goel,
  Hannemann, Motlicek, Qian, Schwarz, Silovsky, Stemmer, and Vesely]{kaldi}
Daniel Povey, Arnab Ghoshal, Gilles Boulianne, Lukas Burget, Ondrej Glembek,
  Nagendra Goel, Mirko Hannemann, Petr Motlicek, Yanmin Qian, Petr Schwarz, Jan
  Silovsky, Georg Stemmer, and Karel Vesely.
\newblock The {K}aldi speech recognition toolkit.
\newblock In \emph{Workshop on Automatic Speech Recognition and Understanding
  (ASRU)}. IEEE, 2011.

\bibitem[Hara et~al.(2018)Hara, Kataoka, and Satoh]{hara3dcnns}
Kensho Hara, Hirokatsu Kataoka, and Yutaka Satoh.
\newblock Can spatiotemporal 3d {CNN}s retrace the history of 2d cnns and
  imagenet?
\newblock In \emph{Proceedings of the Conference on Computer Vision and Pattern
  Recognition(CVPR)}. IEEE, 2018.

\bibitem[Kudo()]{sentencepiece}
Taku Kudo.
\newblock {SentencePiece}: {A} simple and language independent subword
  tokenizer and detokenizer for neural text processing.
\newblock In \emph{Proceedings of the 2018 Conference on Empirical Methods in
  Natural Language Processing (EMNLP)}. ACL.

\bibitem[Chan et~al.(2016)Chan, Jaitly, Le, and Vinyals]{las}
William Chan, Navdeep Jaitly, Quoc~V. Le, and Oriol Vinyals.
\newblock Listen, attend and spell: {A} neural network for large vocabulary
  conversational speech recognition.
\newblock In \emph{Proceedings of the International Conference on Acoustics,
  Speech and Signal Processing (ICASSP)}. IEEE, 2016.

\bibitem[Sennrich et~al.(2017)Sennrich, Firat, Cho, Birch-Mayne, Haddow,
  Hitschler, Junczys-Dowmunt, Läubli, {Miceli Barone}, Mokry, and
  Nadejde]{nematus}
Rico Sennrich, Orhan Firat, Kyunghyun Cho, Alexandra Birch-Mayne, Barry Haddow,
  Julian Hitschler, Marcin Junczys-Dowmunt, Samuel Läubli, Antonio {Miceli
  Barone}, Jozef Mokry, and Maria Nadejde.
\newblock Nematus: a toolkit for neural machine translation.
\newblock In \emph{Proceedings of the European Chapter of the Association for
  Computational Linguistics (EACL). Software Demonstrations}. ACL, 2017.

\bibitem[Bahdanau et~al.(2014)Bahdanau, Cho, and Bengio]{Bahdanau2014}
Dzmitry Bahdanau, Kyunghyun Cho, and Yoshua Bengio.
\newblock Neural machine translation by jointly learning to align and
  translate.
\newblock \emph{Computing Research Repository (CoRR)}, 2014.

\bibitem[Press and Wolf(2017)]{press2016using}
Ofir Press and Lior Wolf.
\newblock Using the output embedding to improve language models.
\newblock In \emph{Proceedings of the Conference of the European Chapter of the
  Association for Computational Linguistics (EACL)}. ACL, 2017.

\bibitem[Srivastava et~al.(2014)Srivastava, Hinton, Krizhevsky, Sutskever, and
  Salakhutdinov]{srivastava2014dropout}
Nitish Srivastava, Geoffrey Hinton, Alex Krizhevsky, Ilya Sutskever, and Ruslan
  Salakhutdinov.
\newblock Dropout: {A} simple way to prevent neural networks from overfitting.
\newblock \emph{Journal of Machine Learning Research (JMLR)}, 2014.

\bibitem[Kingma and Ba(2014)]{kingma2014adam}
Diederik Kingma and Jimmy Ba.
\newblock Adam: A method for stochastic optimization.
\newblock \emph{Journal of Machine Learning Research (JMLR)}, 2014.

\bibitem[Pascanu et~al.(2013)Pascanu, Mikolov, and
  Bengio]{pascanu2013difficulty}
Razvan Pascanu, Tomas Mikolov, and Yoshua Bengio.
\newblock On the difficulty of training recurrent neural networks.
\newblock In \emph{Proceedings of the International Conference on International
  Conference on Machine Learning (ICML)}. IMIS, 2013.

\bibitem[Papineni et~al.(2002)Papineni, Roukos, Ward, and
  Zhu]{Papineni:2002:acl}
Kishore Papineni, Salim Roukos, Todd Ward, and Wei-Jing Zhu.
\newblock {BLEU}: {A} method for automatic evaluation of machine translation.
\newblock In \emph{Proceedings of the Annual Meeting on Association for
  Computational Linguistics (ACL)}. ACL, 2002.

\bibitem[Lin and Och(2004)]{rouge}
Chin-Yew Lin and Franz~Josef Och.
\newblock Automatic evaluation of machine translation quality using longest
  common subsequence and skip-bigram statistics.
\newblock In \emph{Proceedings of the Meeting of the Association for
  Computational Linguistics (ACL)}. ACL, 2004.

\end{thebibliography}

\newpage

\appendix
\section{Appendix}
\label{sec:appendix}
\subsection{\dataset{} Examples}
\label{sec:how2examples}
\begin{figure}[t]
\vspace*{25px}
\begin{subfigure}[c]{0.3\textwidth}
\includegraphics[width=3.5cm,height=2.5cm]{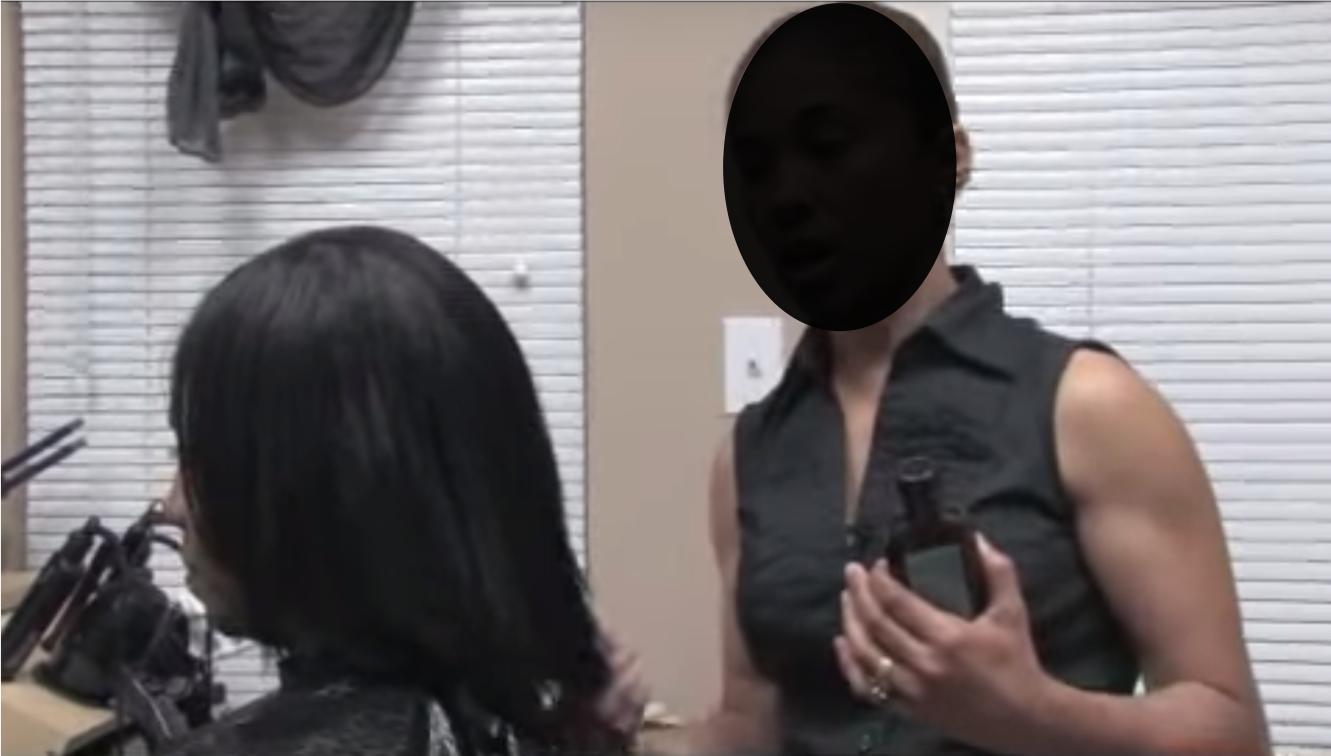}\\
\end{subfigure}
\begin{subfigure}[c]{0.63\textwidth}
\textbf{Actually I use this beautiful moroccan oil and it's really wonderful on the hair.}\\[.1cm]
\textit{Na verdade eu uso este belo óleo marroquino e é realmente maravilhoso no cabelo.}\\[.1cm]
\begin{tikzpicture}
\node [text width=25em, draw, gray, rounded corners, inner sep=1ex] {\textcolor{black}{Relaxed African-American hair should be moisturized and washed to maintain good and healthy hair. Care for African-American, biracial or ethnic hair with tips from a professional hairstylist in this free video on hair care.} 
};
\end{tikzpicture}
\end{subfigure}

\vspace*{25px}

\begin{subfigure}[c]{0.3\textwidth}
\includegraphics[width=3.5cm,height=2.5cm]{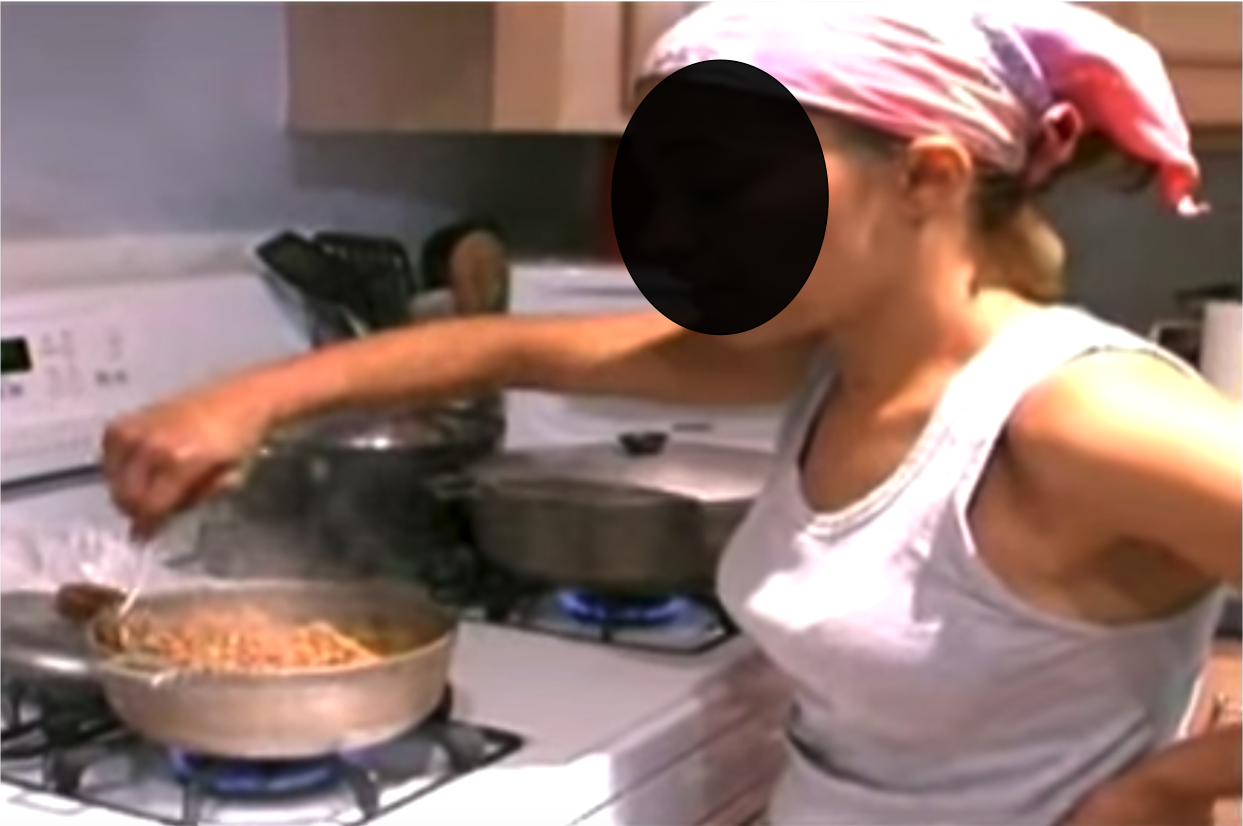}\\
\end{subfigure}
\begin{subfigure}[c]{0.63\textwidth}
\textbf{Like said you can cook this pretty quickly with your family.}\\[.1cm]
\textit{Como disse você pode cozinhar isso muito rapidamente com sua família.}\\[.1cm]
\begin{tikzpicture}
\node [text width=25em, draw, gray, rounded corners, inner sep=1ex] {\textcolor{black}{Learn how to cook and serve picadillo con arroz with expert cooking tips in this free classic Cuban recipe video clip.}
};
\end{tikzpicture}
\end{subfigure}

\vspace*{25px}

\begin{subfigure}[c]{0.3\textwidth}
\includegraphics[width=3.5cm,height=2.5cm]{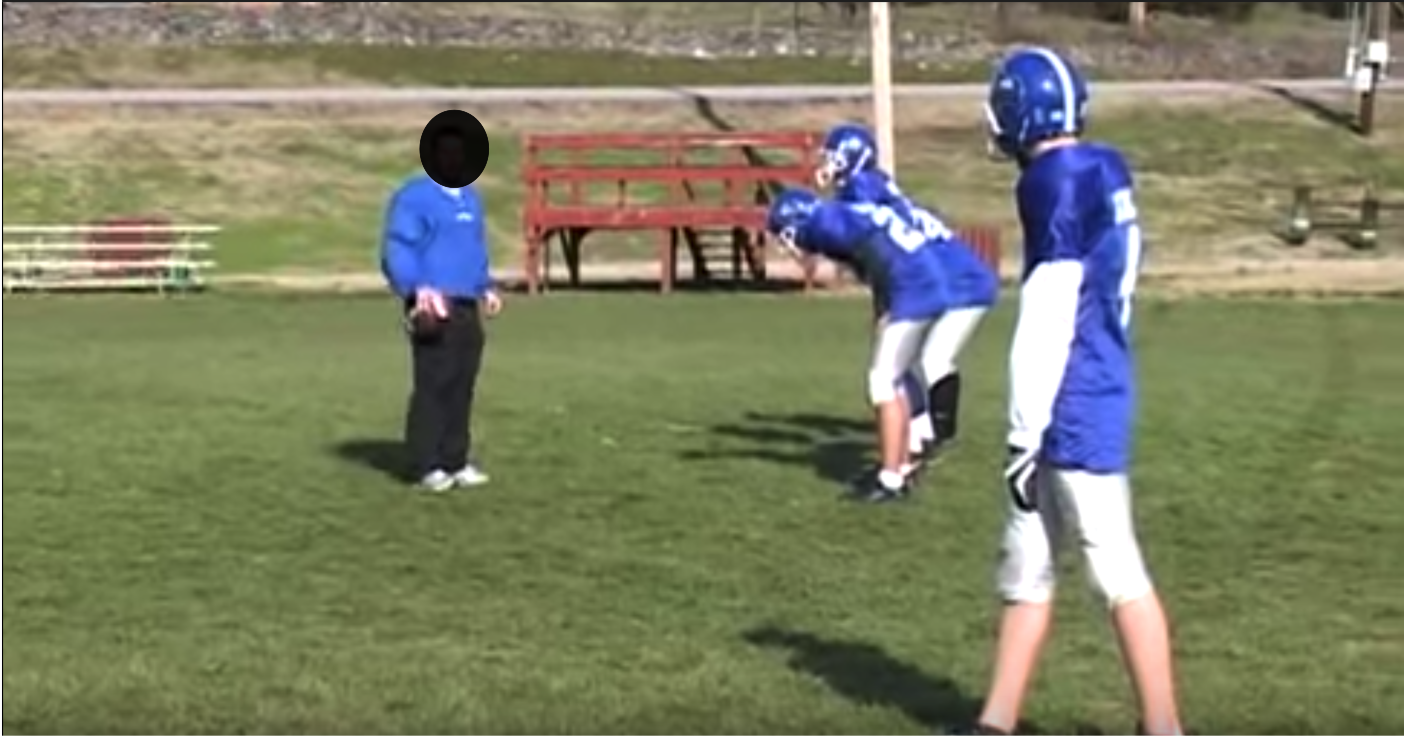}\\
\end{subfigure}
\begin{subfigure}[c]{0.63\textwidth}
\textbf{When your wide receivers get into their stance they want to have one foot forward, they want to have good position and be ready to fire off the line.}\\[.1cm]
\textit{Quando seus receptores largos entram em sua posição eles devem ter um pé para a frent,e eles devem ter uma boa posição e estar pronto para disparar a linha.}\\[.1cm]
\begin{tikzpicture}
\node [text width=25em, draw, gray, rounded corners, inner sep=1ex] {\textcolor{black}{Learn some great tips on how to line up as a receiver in this free video clip on how to play football.}
};
\end{tikzpicture}
\end{subfigure}

\caption{\label{fig:twocol_sample} Examples of the {\dataset} dataset that reflect its sample variability and modality correlation. The image is a randomly selected frame from a concrete segment of the video, the text in \textbf{bold} is the utterance pronounced during this segment, the Portuguese text in \textit{italic} are the translations corresponding to the utterance and finally, the summary of the whole video is placed inside the rectangle.}
\end{figure}

In Figure~\ref{fig:twocol_sample}, we list three typical instances from the {\dataset} dataset. In these examples, we can see the correspondence between the content of the video frame, the summary, and the utterance. This multimodal correspondence is what systems can exploit by using the {\dataset} dataset. 
In the first example, a hairdresser is explaining how to use a specific hair product. In that case, the visual elements (\textit{i.e.}, hair product, hairdresser) and the scene, a hairdressing salon, are a rich source of context. In the second example, a woman is cooking in a kitchen with many cooking devices. In the third example, we can infer the acoustic environment (\textit{i.e.}, outdoors) by the scene. 

\subsection{Modality Alignment and Data Checks}
\label{sec:audio_seg}

To combine all modalities (\textit{i.e.}, speech, video, transcriptions, and translations) successfully, we need to establish and validate their correspondence in time. While the audio transcription is generated from subtitles, these do not always correspond to the actual audio. We thus decided to generate token-level (e.g. word-level) time stamps that link text, audio, and video modality. From these, utterance-level start and end times were also calculated.

To align text and audio, 
we perform a Viterbi alignment between the transcriptions, which were provided by the users who uploaded the videos, and the audio track of {\dataset}, using Kaldi's~\cite{kaldi} Wall Street Journal (WSJ) GMM/ HMM acoustic model. This alignment process estimates the start and end times of each sentence in the audio track. Finally, by using the estimated alignments, we can segment the audio and video track according to the utterances.


To make sure the data will be suitable for the proposed use, we validated two properties: \textbf{First}, we verified that the word alignment is indeed accurate by manually inspecting randomly chosen utterances. The 300\,h sub-set was selected to give good alignment scores, and the WSJ model seemed to perform best in that respect: ``good'' (according to the score) utterances were indeed accurately aligned, when compared to alignments generated with other models, including those developed for speech synthesis. \textbf{Second}, the ``transcription'' data has been generated from video subtitles, which were not meant to be verbatim and highly accurate ``transliterations'' of the spoken content. Rather, the ``transcription'' text is a somewhat canonical form of the spoken word, which is fine for our proposed uses, although it may lead to slightly higher overall word error rates for the speech-to-text tasks.








\subsection{Feature Extraction and Processing}

\paragraph{Speech Features}
We used Kaldi \cite{kaldi} to extract 40-dimensional filter bank features from \textit{16kHz} raw speech using a time window of \textit{25ms} with \textit{10ms} frame shift and concatenated 3-dimensional pitch features to obtain the final 43-dimensional speech features. A \textit{per-video} Cepstral Mean and Variance Normalization (CMVN) is further applied to account for speaker variability.

\paragraph{Visual Features}
A 2048-dimensional feature vector per 16 frames is extracted from the videos using a CNN trained to recognize 400 different actions \cite{hara3dcnns}. It should be noted that this results in a sequence of feature vectors per video rather than a single/global one. In order to obtain the latter, we average pooled the extracted features into a single 2048-dimensional feature vector which will represent all sentences segmented out of a single video.

\paragraph{Text Features}
All texts are normalized, lowercased and filtered from punctuation. A SentencePiece model \cite{sentencepiece} is learned separately for English and Portuguese resulting into vocabularies of 5K each except for summarization which uses word-level tokens.

\subsection{Architecture Details}
\label{sec:arch_details}
\paragraph{Automatic Speech Recognition} 
We use a 6-layer pyramidal encoder \cite{las} (with interleaved $tanh$ projection layers) where the middle two layers skip every other input resulting into a subsampling rate of 4. The decoder is a 2-layer conditional GRU (CGRU) decoder \cite{nematus}, a transitive decoder where the hidden state of the second GRU is fed back to the first GRU. The first GRU is initialized with the mean encoder hidden state transformed using a $tanh$ layer. A feed-forward attention mechanism \cite{Bahdanau2014} is used inside the decoder and the input and output embeddings are tied \cite{press2016using}.

\paragraph{Multimodal Automatic Speech Recognition} 
For  multimodal ASR, we apply video adaptive training \cite{context_mmasr,palaskar_mmasr}
with a learned linear transformation of the visual feature vector $v$ that will act as a \textit{visual bias} over a given speech feature $x_t$ at time $t$. The shifted speech feature $\hat{x_{t}}$ is thus computed as follows:
\begin{gather*}
\hat{x_{t}} = x_t + \left(\mathbf{W}^{T} v + b\right)\label{asr:linear}
\end{gather*}
To train this model, we first initialize the model parameters using a previously trained ASR and jointly optimize all parameters including $\mathbf{W}$ and $b$ above.

\paragraph{Machine Translation}
We train a sequence-to-sequence neural MT model with a 2-layer bidirectional GRU \cite{cho2014gru} encoder and a 2-layer conditional GRU decoder\cite{nematus}. A dropout \cite{srivastava2014dropout} with $0.3$ drop probability is used after source embeddings, source encodings and before the final $softmax$ operation.

\paragraph{Multimodal Summarization}
We follow the hierarchical attention approach~\cite{libovicky2017attention} to combine textual and visual modalities. Unlike previous multimodal ASR, MT and STT architectures, the visual features described in Section~\ref{sec:dataset} are now used as-is, i.e. as a sequence of 2048-dimensional vectors rather than being average pooled into a single vector.

\subsection{Training Details}
\label{sec:train_details}

Unless otherwise specified, we use ADAM \cite{kingma2014adam} as the optimizer with an initial learning rate of 0.0004. The gradients are clipped to have unit norm \cite{pascanu2013difficulty}. The training is early stopped if the task performance on validation set does not improve for 10 consecutive epochs. Task performance is assessed using Word Error Rate (WER) for speech recognition, BLEU \cite{Papineni:2002:acl} for translation tasks and ROUGE-L \cite{rouge} for the summarization task. The learning rate is halved whenever the task performance does not improve for two consecutive epochs. All systems are trained three times with different random initializations. The hypotheses are decoded using beam search with a beam size of 10. We report the average results of the three runs. We use \textit{nmtpytorch}~\cite{nmtpy2017} to train the models.

\end{document}